%% file: root.tex
\PassOptionsToPackage{draft}{hyperref}
\documentclass[a4paper, 10 pt, conference]{ieeeconf}  % Comment this line out if you need a4paper

\IEEEoverridecommandlockouts                              % This command is only needed if 
\usepackage{balance}
\usepackage{lipsum}
\usepackage{subcaption}
\usepackage{amssymb}
\usepackage[nolist]{acronym} 
\usepackage{tikz,pgfplots}
\usepackage{bm}
\usepackage[noadjust]{cite}
\usepackage{amsmath}

\usepackage[bookmarks=false]{hyperref}
\usepackage{textcomp}
\usepackage{float}
\usepackage{algorithm}
\usepackage{tikz}   %TikZ is required for this to work.  Make sure this exists before the next line
\usepackage{3dplot} %requires 3dplot.sty to be in same directory, or in your LaTeX installation
\usepackage{algpseudocode}
\usepackage[all]{comment} % Highlight edits and show comments
\usepackage{amsmath,amsfonts,amssymb}
\usepackage{mathtools}
\usepackage{multirow}
\usepackage{bm}
\usepackage[T1]{fontenc}
\usepackage{esvect}

\usepackage[font=small,labelfont=bf]{caption}

\usepackage{array}
\newcommand\ChangeRT[1]{\noalign{\hrule height #1}}

\usetikzlibrary{arrows.meta, 
                bending, 
                calc, chains, 
                decorations.pathmorphing,  % added
                positioning,patterns,decorations.markings
                }

\tikzset{pill/.style={minimum width=1.2cm,minimum height=6mm,rounded
corners=3mm,draw},
reactor/.style={circle,draw,minimum size=6mm,path picture={
\draw (-3mm,0) -- (3mm,0) (0,-3mm) -- (0,3mm);
\fill (0,0) -- (3mm,0) arc(0:-90:3mm) -- cycle;
\fill (0,0) -- (-3mm,0) arc(180:90:3mm) -- cycle;
}}}

\title{\LARGE \bf
Robust Dynamic Walking for a 3D Dual-SLIP Model under One-Step Unilateral Stiffness Perturbations: Towards Bipedal Locomotion over Compliant Terrain}
\author{Chrysostomos Karakasis, \textit{IEEE Student Member}, Ioannis Poulakakis, \textit{IEEE Senior Member}, \\and Panagiotis Artemiadis$^*$, \textit{IEEE Senior Member}%
\thanks{*This material is based upon work supported by the National Science Foundation under Grants No. \#2020009, \#2015786, \#2025797, and \#2018905. This scientific paper is partially supported by the Onassis Foundation - Scholarship ID: F ZQ029-1/2020-2021.}% <-this % stops a space
\thanks{Chrysostomos Karakasis, Ioannis Poulakakis, and Panagiotis Artemiadis are with the Mechanical Engineering Department, at the University of Delaware, Newark, DE 19716, USA.
        {\tt\small \{chryskar, poulakas, partem\}@udel.edu}}%
\thanks{$^*$Corrresponding author: partem@udel.edu}%
}

\begin{document}

\maketitle
\thispagestyle{empty}
\pagestyle{empty}

\input{acro_list}
\setlength{\textfloatsep}{10pt plus 2.0pt minus 4pt}
%%%%%%%%%%%%%%%%%%%%%%%%%%%%%%%%%%%%%%%%%%%%%%%%%%%%%%%%%%%%%%%%%%%%%%%%%%%%%%%%
\begin{abstract}
Bipedal walking is one of the most important hallmarks of human that robots have been trying to mimic for many decades. Although previous control methodologies have achieved robot walking on some terrains, there is a need for a framework allowing stable and robust locomotion over a wide range of compliant surfaces. This work proposes a novel biomechanics-inspired controller that adjusts the stiffness of the legs in support for robust and dynamic bipedal locomotion over compliant terrains. First, the 3D Dual-SLIP model is extended to support for the first time locomotion over compliant surfaces with variable stiffness and damping parameters. Then, the proposed controller is compared to a Linear-Quadratic Regulator (LQR) controller, in terms of robustness on stepping on soft terrain. The LQR controller is shown to be robust only up to a moderate ground stiffness level of 200 $kN/m$, while it fails in lower stiffness levels. On the contrary, the proposed controller can produce stable gait in stiffness levels as low as 30 $kN/m$, which results in a vertical ground penetration of the leg that is deeper than 10\% of its rest length. The proposed framework could advance the field of bipedal walking, by generating stable walking trajectories for a wide range of compliant terrains useful for the control of bipeds and humanoids, as well as by improving controllers for prosthetic devices with tunable stiffness.

\end{abstract}

%%%%%%%%%%%%%%%%%%%%%%%%%%%%%%%%%%%%%%%%%%%%%%%%%%%%%%%%%%%%%%%%%%%%%%%%%%%%%%%%
\input{introduction}

\input{methods}

\input{results}

\input{conclusions}

\bibliographystyle{IEEEtran}
\bibliography{mybib}

\end{document}

%% file: acro_list.tex
\begin{acronym}[TD] % Give the longest label here so that the list is nicely aligned
\acro{MS}{Midstance}
\acro{TD}{Touchdown}
\acro{LH}{Lowest Height}
\acro{LO}{Lift Off}
\acro{GRF}{ground reaction force}
\acro{CoM}{Center of Mass}
\acro{LQR}{Linear-Quadratic Regulator}
\acro{SS}{single support}
\acro{DS}{double support}
\acro{Dual-SLIP}{Dual Spring-Loaded Inverted Pendulum}
\acro{HC}{Hunt-Crossley}
\acro{ULIP}{Unilateral Low Impedance Perturbations}
\end{acronym}

%% file: introduction.tex
\section{INTRODUCTION}
%1st State problem: Bipeds and Humanoids on compliant terrain
%2nd Why modelling and 3D Dual-SLIP
%3rd Why compliance - VST and vasilopoulos
%4th Why biomechanics 

Although bipedal robots have evolved drastically over the years, most proposed frameworks and controllers have been designed and tested only over rigid surfaces \cite{venancio2020terrain}. However, in real-life situations these systems are faced with motion tasks over unpredictable non-rigid terrains with highly variable ground parameters, such as stiffness and damping \cite{wang2020situ}. Therefore, there is a need for a framework that takes into account the ground properties and allows for robust and stable locomotion over variable impedance terrain. 
% impedance = stiffness + damping

%%%%%%%%%%%%%%%%%%%%%%%%%%
% Figure from methods placed on the front page
\input{compliant_3d_dual_slip}   
%%%%%%%%%%%%%%%%%%%%%%%%%%

%%%% PANOS MOVED THIS PARAGRAPH HERE
% Previous work on biped locomotion over compliant terrain. 
Previous work on bipedal robotic locomotion over compliant terrain can be divided into two groups. On one hand, researchers have treated terrain compliance as an external disturbance used to verify the robustness of their control approaches \cite{mesesan2019dynamic,hopkins2016optimization}. However, the alleged robustness is limited, since only specific terrains have been tested (grass, gravel and gym mattress). On the other hand, researchers have developed methods for estimating terrain properties aiming to apply terrain-specific feedback control techniques \cite{wang2020situ,venancio2020terrain}. As a result, there has not been any control approach for bipeds based on a ground model that provides robustness over a wide range of compliant surfaces.

%why 3D Dual-SLIP model
To this day, considerable research effort has focused on using simple models as ``templates'' for the study of complex dynamical systems, such as monopods, bipeds and humanoids, as well as humans~\cite{full1999templates,vasilopoulos2014compliant,poulakakis2009spring}. For modeling human walking, Geyer et al. concluded that the 2D \acf{Dual-SLIP} is a more accurate model than other simple models, as it was able to produce human-like \ac{CoM} vertical oscillations and \ac{GRF} responses \cite{geyer2006compliant}. Recently, the model was extended to three dimensions (3D \ac{Dual-SLIP}) to capture the \ac{CoM} lateral sway observed in human walking \cite{liu2015dynamic}. Moreover, control methodologies have been proposed based on actuated versions of the 3D \ac{Dual-SLIP} for humanoid locomotion over rigid, uneven and rough terrain \cite{liu2015trajectory,xiong2021slip}. However, to the best of our knowledge, the 3D Dual-SLIP has never been implemented on compliant terrain.

% % Previous works showing importance of leg stiffness
%Background on leg stiffness research in biomechanics and robotics
%why use leg stiffness to handle perturbations
Previous research in legged systems has indicated that leg stiffness is crucial for regulating external disturbances and improving energy efficiency \cite{visser2012robust,liu2018switchable}. Furthermore, research in biomechanics has shown that humans increased leg stiffness as surface stiffness decreased \cite{ferris1998running,ferris1999runners}. Those findings support the idea that adjusting the leg stiffness can provide robustness and versatility during locomotion over rigid and compliant terrains.

% Summarize our work 
This work proposes a novel controller for robust and dynamic bipedal locomotion over compliant terrain. First, the 3D \ac{Dual-SLIP} model is extended to support for the first time locomotion over compliant surfaces with variable stiffness and damping parameters, using the \ac{HC} model. A nonlinear optimization approach and a \ac{LQR} controller similar to those proposed for rigid terrains (\cite{liu2015dynamic}) are implemented to achieve periodic walking gaits over stiff terrains. However, the \ac{LQR} controller is shown to be inadequate for stable walking over a moderate ground stiffness level of 200 $kN/m$ or less. For this reason, a new biomechanics-inspired controller is introduced that adjusts the stiffness of the legs in support. The proposed controller is tested on very soft terrains and results in stable walking after one-step unilateral stiffness perturbations at stiffness levels as low as 30 $kN/m$, which resembles the stiffness of a foam pad. It is shown that on such soft terrains, the leg sinks into the soft ground up to 11.49 $cm$, which is significant for the rest length of the legs (1 $m$). Despite that, the proposed framework achieves stable walking and a fast recovery (less than 10 steps) after the 1-step perturbation. As a result, robust dynamic walking over extremely low one-step unilateral stiffness perturbations can be achieved using the proposed controller. The proposed framework could advance the field of bipedal walking, by generating stable walking trajectories for various compliant terrains useful for the control of bipeds and humanoids, as well as by improving controllers for prosthetic devices with tunable stiffness.

%% file: compliant_3d_dual_slip.tex
%Angle Definitions
%-----------------
\tikzset{>=latex} % for LaTeX arrow head

%set the plot display orientation
%synatax: \tdplotsetdisplay{\theta_d}{\phi_d}
% \tdplotsetmaincoords{60}{60}
\tdplotsetmaincoords{60}{60}

%define polar coordinates for some vector
%TODO: look into using 3d spherical coordinate system
\pgfmathsetmacro{\rvec}{.89}
\pgfmathsetmacro{\thetavec}{35}
\pgfmathsetmacro{\phivec}{50}

%start tikz picture, and use the tdplot_main_coords style to implement the display 
%coordinate transformation provided by 3dplot
\begin{figure}[t]
    \centering
\begin{tikzpicture}[scale=5, every node/.style={scale=1},tdplot_main_coords,spring/.style = {thick, decorate,       % new, 
                 decoration={zigzag,pre length=3mm,post length=3mm,segment length=10}
}]
% \tikzstyle{damper}=[thick,tdplot_main_coords,decoration={markings,
%   mark connection node=dmp,
%   mark=at position 0.5 with
%   {
%     \node (dmp) [thick,inner sep=0pt,tdplot_main_coords,transform shape,rotate=-90,minimum width=5pt,minimum height=2pt,draw=none] {};
%     \node (dmp_ne) [tdplot_main_coords] at ($(dmp)+(0,5pt,0.0pt)$) {};
%     \draw [thick,tdplot_main_coords] (dmp_ne) -- (dmp.south east) -- (dmp.south west) -- ($(dmp.north west)+(10pt,0)$);
%     \draw [thick,tdplot_main_coords] ($(dmp.north)+(0pt,-2pt)$) -- ($(dmp.north)+(0pt,2pt)$);
%   }
% }, decorate]
% \tikzstyle{ground}=[fill,pattern=north east lines,draw=none,minimum width=0.75cm,minimum height=0.3cm]
\tikzstyle{ground}=[fill,tdplot_main_coords,rotate = 0,canvas is yz plane at x=0, pattern=north east lines,draw=none,minimum width=0.5cm,minimum height=0.2cm]

%set up some coordinates 
%-----------------------

\coordinate (O) at (0,0,0);
\draw[thin] (-0.1,0.1,0) -- (-0.1,0.1,0) node[xshift=-2.5em,yshift=-0.5em] (pa){$\bm{p}_{f,A}$};

%determine a coordinate (P) using (r,\theta,\phi) coordinates.  This command
%also determines (Pxy), (Pxz), and (Pyz): the xy-, xz-, and yz-projections
%of the point (P).
%syntax: \tdplotsetcoord{Coordinate name without parentheses}{r}{\theta}{\phi}
\tdplotsetcoord{P}{\rvec}{\thetavec}{\phivec}
\tdplotsetcoord{PB}{0.9}{90}{50}
\tdplotsetcoord{P_2}{0.54}{\thetavec-1}{\phivec}
\tdplotsetcoord{PB_2}{0.9}{50}{36.5}
\coordinate (T) at (P);
%draw figure contents
%--------------------
\draw[color=white] (PB) - ++(-0.25,0.1,-0.17) node[color=black,yshift=3pt, anchor=north east] (pb){$\bm{p}_{f,B}$};

%draw the main coordinate system axes
\draw[thick,->] (-0.1,0.1,0) -- (0.2,0.1,0) node[xshift=0.4em,yshift=0em]{$\hat{\bm{x}}$};
\draw[thick,->] (-0.1,0.1,0) -- (-0.1,0.3,0) node[xshift=0.5em,yshift=0em]{$\hat{\bm{y}}$};
\draw[thick,->] (-0.1,0.1,0) -- (-0.1,0.1,0.125) node[anchor=east]{$\hat{\bm{z}}$};
\draw[->] (-0.1,0.1,0) -- ++(0,0,0.2) node[anchor=south]{$\bm{{F}}_{g,A}$};

\draw[thick,->] (PB) -- ++(0.2,0,0) node[xshift=0.3em,yshift=-0.15em]{$\hat{\bm{x}}$};
\draw[thick,->] (PB) -- ++(0,0.2,0) node[anchor=west]{$\hat{\bm{y}}$};
\draw[thick,->] (PB) -- ++(0,0,0.125) node[anchor=west]{$\hat{\bm{z}}$};
\draw[->] (PB) -- ++(0,0,0.2) node[yshift=0.7em]{$\bm{{F}}_{g,B}$};

%draw a vector from origin to point (P) 
\draw[-,color=red,spring] (-0.1,0.1,0) -- (P);
% \draw[-,color=red,spring] (-0.34,0.43,0) -- (-0.605,0.8,0);
% \draw[-,color=red,thick] (-0.1,0.1,0) -- (-0.34,0.43,0);
% \draw[-,color=red,thick] (-0.605,0.8,0) -- (P);
\node[align=center,anchor=south east] at (P_2) {\textcolor{red}{$\bm{l}_{A}$}};
\node[align=center,anchor=north west] at (P_2) {\textcolor{red}{$k_{A}$}};
\draw[-,color=blue,spring] (P) -- (PB);
% \draw[-,color=blue,spring] (P) ++(0.4,-0.10,0) -- ++(0.5,-0.125,0)(PB);
\node[align=center] at (PB_2) [xshift=-0.2em,yshift=-1.8em] {\textcolor{blue}{$k_{B}$}};
\node[align=center] at (PB_2) [xshift=2em,yshift=-0.8em] {\textcolor{blue}{$\bm{l}_{B}$}};
% \draw[-,color=blue,thick] (P) -- ++(0.4,-0.10,0);
% \draw[-,color=blue,thick] (P) ++(0.4,-0.10,0) ++(0.5,-0.125,0) -- (PB);

\draw[thin] (P) -- (P) node[pos=1,reactor,scale=1] (M){~};
\draw[thin] (PB) -- (PB) node[pos=1,reactor,scale=0.5] (M2){~};
\draw[thin] (-0.1,0.1,0) -- (-0.1,0.1,0) node[pos=1,reactor,scale=0.5] (M1){~};
% \draw[thin] (P) -- (PB) node[pos=1,scale=0.5] (PBF){~};
\draw[dashed] (Pxy) -- ([xshift=1.35em,yshift=-0.1em]PB) node[pos=1,scale=0.5] (PBD){~};
\draw[thick,->] (Pxy) -- ++(0.2,0,0) node[anchor=north east]{$\hat{\bm{x}}$};

\draw[thin] (M) -- (M) node[xshift=1.5em,yshift=0em] (tf){$m$};
\draw[thin] (M2) -- (M2) node[xshift=-2em,yshift=0.7em] (tf2){$m_{f,B}$};
\draw[thin] (M1) -- (M1) node[xshift=-2.7em,yshift=0.7em] (tf3){$m_{f,A}$};
\draw[thin] (M) -- ([xshift=-0.5pt]M) node[xshift=-1.5em] (pc){$\bm{p}_{c}$};
\draw[->] (P) -- ([xshift=4.5pt,yshift=5pt]P) node[anchor=west] (Fl1){$\bm{{F}}_{s,A}$};
\draw[->] (-0.1,0.1,0) -- (+0.11,-0.2,0) node[anchor=north,xshift=-1.5em,yshift=1em] (Fl5){$\bm{{F}}_{s,A}$};
% \draw[thick] (0,0,0)  -- (P) node[pos=1,sloped,reactor,scale=0.5] (M){~};

%draw projection on xy plane, and a connecting line
\draw[dashed] (-0.1,0.1,0) -- (Pxy);
\draw[dashed] (P) -- (Pxy);
\draw[->] (P) -- ([yshift=11pt]Pxy) node[anchor=west] (g){$\bm{g}$};
\draw[->] (PB) -- ([yshift=-5pt]PB) node[xshift=-4pt,anchor=east] (g2){$\bm{g}$};
\draw[->] (M1) -- ($(M1)+(0,0,-0.2pt)$) node[anchor=west,xshift=0.5em,yshift=0.2em] (g3){$\bm{g}$};
\draw[->] (P) -- ([xshift=-3pt,yshift=5pt]P) node[anchor=east] (Fl1){$\bm{{F}}_{s,B}$};
\draw[->] (PB) -- ([xshift=3pt,yshift=-5pt]PB) node[anchor=west] (Fl2){$\bm{{F}}_{s,B}$};

%draw the angle \phi, and label it
%syntax: \tdplotdrawarc[coordinate frame, draw options]{center point}{r}{angle}{label options}{label}
% \tdplotdrawarc[->]{(-0.1,0.1,0)}{0.2}{0}{\phivec-15}{xshift=1em,yshift=-0.5em}{$\theta_{1}$}
\tdplotdrawarc[->]{(Pxy)}{0.1}{0}{50}{yshift=0.5em,anchor=north}{\hspace{1.5em}$\phi$}
\tdplotdrawarc[->]{(PB)}{0.1}{0}{50}{xshift=0.4em,yshift=0.5em,anchor=north}{\hspace{1.5em}$\phi$}

%set the rotated coordinate system so the x'-y' plane lies within the
%"theta plane" of the main coordinate system
%syntax: \tdplotsetthetaplanecoords{\phi}
\tdplotsetthetaplanecoords{60}
% \pgfmathsetmacro{\rvec}{.8}
% \pgfmathsetmacro{\thetavec}{60}
% \pgfmathsetmacro{\phivec}{70}
%draw theta arc and label, using rotated coordinate system
% \tdplotdrawarc[tdplot_rotated_coords,thick,<-]{(0,0,0)}{0.3}{34}{90}{anchor=west}{$\theta_{2}$}
\tdplotdrawarc[tdplot_rotated_coords,thick,<-]{(PB)}{0.35}{-25}{95}{anchor=south}{$\theta$}

% %draw some dashed arcs, demonstrating direct arc drawing
% \draw[dashed,tdplot_rotated_coords] (\rvec,0,0) arc (0:90:\rvec);
% \draw[dashed] (\rvec,0,0) arc (0:90:\rvec);

%set the rotated coordinate definition within display using a translation
%coordinate and Euler angles in the "z(\alpha)y(\beta)z(\gamma)" euler rotation convention
%syntax: \tdplotsetrotatedcoords{\alpha}{\beta}{\gamma}
\tdplotsetrotatedcoords{\phivec}{\thetavec}{0}

%%%%%%%%%%%%%%%%%%%%%%%%%%%%%%%%%%%%%%%%%%%%%%%%%%%%%%%%%%%%%%%%%%%%%%%%%%%%%
\node (g1) [ground, rotate=0, minimum width=1cm] at ($(M1)+(0,0,-0.5pt)$) {};
% \draw[thick,tdplot_main_coords] ($(g1)+(0,0.05pt,0.05pt)$) -- ($(M1)+(0,0.05pt,-0.05pt)$);
% \draw ($(g1)+(0,0.05pt,0.05pt)$) -- ($(M1)+(0,0.05pt,-0.05pt)$);
\draw[thick,opacity=0.5] ($(g1)+(0,0.05pt,0.2pt)$) -- ($(M1)+(0,0.05pt,-0.1pt)$);
\node (u2) at ($(g1)+(0,0.05pt,0.2pt)$) {};
\draw[thick,opacity=0.5] ($(u2)+(0,-0.02pt,0)$) -- ($(u2)+(0,0.02pt,0)$);
\draw[thick,opacity=0.5] ($(g1)+(0,0.05pt,0.0pt)$) -- ($(g1)+(0,0.05pt,0.15pt)$);
\node (u3) at ($(g1)+(0,0.05pt,0.15pt)$) {};
\draw[thick,opacity=0.5] ($(u3)+(0,-0.03pt,0)$) -- ($(u3)+(0,0.03pt,0)$);
\draw[thick,opacity=0.5] ($(u3)+(0,-0.03pt,0)$) -- ($(u3)+(0,-0.03pt,0.1pt)$);
\draw[thick,opacity=0.5] ($(u3)+(0,+0.03pt,0)$) -- ($(u3)+(0,+0.03pt,0.1pt)$);
\draw[spring,segment length=6,opacity=0.5] ($(g1)+(0,-0.05pt,0pt)$) -- ($(M1)+(0,-0.05pt,-0.1pt)$);
\draw[thick,opacity=0.5] ($(M1)+(0,-0.05pt,-0.1pt)$) -- ($(M1)+(0,0.05pt,-0.1pt)$);
\draw[thick,opacity=0.5] (M1) -- ($(M1)+(0,0pt,-0.1pt)$);
\node[opacity=0.5] (u4) at ($(g1)+(0,0.2pt,0.15pt)$) {$b_{g,A}$};
\node[opacity=0.5] (u5) at ($(g1)+(0,-0.17pt,0.24pt)$) {$k_{g,A}$};
%=====================================================
\node (g2) [ground, rotate=0, minimum width=1cm] at ($(M2)+(0,0,-0.5pt)$) {};
\draw[thick,opacity=0.5] ($(g2)+(0,0.05pt,0.2pt)$) -- ($(M2)+(0,0.05pt,-0.1pt)$);
\node (u2) at ($(g2)+(0,0.05pt,0.2pt)$) {};
\draw[thick,opacity=0.5] ($(u2)+(0,-0.02pt,0)$) -- ($(u2)+(0,0.02pt,0)$);
\draw[thick,opacity=0.5] ($(g2)+(0,0.05pt,0.0pt)$) -- ($(g2)+(0,0.05pt,0.15pt)$);
\node (u3) at ($(g2)+(0,0.05pt,0.15pt)$) {};
\draw[thick,opacity=0.5] ($(u3)+(0,-0.03pt,0)$) -- ($(u3)+(0,0.03pt,0)$);
\draw[thick,opacity=0.5] ($(u3)+(0,-0.03pt,0)$) -- ($(u3)+(0,-0.03pt,0.1pt)$);
\draw[thick,opacity=0.5] ($(u3)+(0,+0.03pt,0)$) -- ($(u3)+(0,+0.03pt,0.1pt)$);
\draw[spring,segment length=6,opacity=0.5] ($(g2)+(0,-0.05pt,0pt)$) -- ($(M2)+(0,-0.05pt,-0.1pt)$);
\draw[thick,opacity=0.5] ($(M2)+(0,-0.05pt,-0.1pt)$) -- ($(M2)+(0,0.05pt,-0.1pt)$);
\draw[thick,opacity=0.5] (M2) -- ($(M2)+(0,0pt,-0.1pt)$);
\node[opacity=0.5] (u4) at ($(g2)+(0,0.2pt,0.15pt)$) {$b_{g,B}$};
\node[opacity=0.5] (u5) at ($(g2)+(0,-0.17pt,0.24pt)$) {$k_{g,B}$};
%%%%%%%%%%%%%%%%%%%%%%%%%%%%%%%%%%%%%%%%%%%%%%%%%%%%%%%%%%%%%%%%%%%%%%%%%%%%%

\end{tikzpicture}
    \caption{The extended 3D \ac{Dual-SLIP} model in compliant terrain during \ac{DS} at Touchdown.}
    \label{fig::DS_diagram_compliant}
\end{figure}
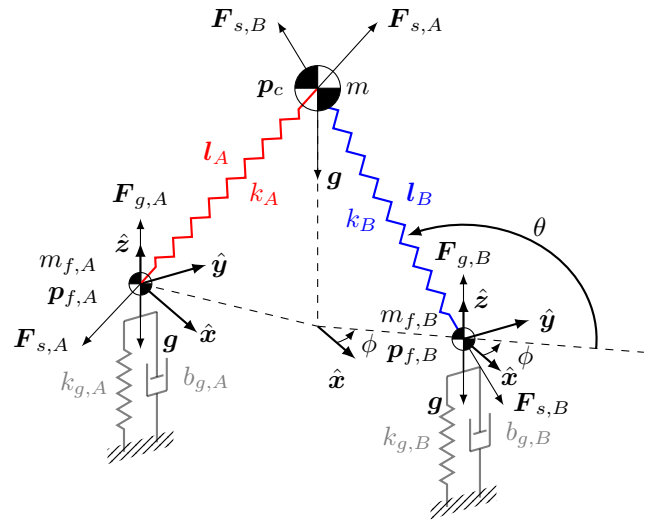

%% file: methods.tex
\section{Methods}
\label{section_methods}

% Provide summary of this section
In order to address biped locomotion over compliant terrain, we first extend the biped walking model 3D \ac{Dual-SLIP}, previously proposed for rigid terrain, to support locomotion over compliant surfaces. Then, we analyze the methodology for finding periodic gaits in such surfaces and achieving them using a standard feedback controller. Finally, we define the induced lower stiffness perturbations and introduce the proposed biomechanics-inspired controller.
%%%%%%%%%%%%%%%%%%%%%%%%%%%%%%%%%%%%%%%%%%%%%%%%%%%%%%%%%%%%%%%%%%%%%%%%%%%%%%%%%%%%%%%%%%%%%%%%%%%%%%%%%%%%%%%%%%%
\subsection{The 3D \ac{Dual-SLIP} Model on Compliant Terrain}
\label{subsection_3D_Dual_SLIP}

The 3D \ac{Dual-SLIP} model was introduced in~\cite{liu2015dynamic}, as an extension of the 2D \ac{Dual-SLIP} model~\cite{geyer2006compliant}, in order to capture both the lateral sway and vertical oscillations of the \ac{CoM} observed in human walking. For brevity, only the extended 3D \ac{Dual-SLIP} will be presented here, as the regular model proposed for locomotion over rigid terrain has been analyzed in depth in previous works~\cite{liu2015dynamic,liu2015dual}.

The model consists of a point mass with two spring legs attached to it, as shown in Fig. \ref{fig::DS_diagram_compliant}. Adopting the notation of~\cite{liu2015dynamic}, $m>0$ is the point mass, and $\bm{p}_{c}= \begin{bmatrix} x_{c} & y_{c} & z_{c}\end{bmatrix}^{\intercal} \in \mathbb{R}^3$ denotes the \ac{CoM} position with respect to an inertial frame of reference. For each leg $i \in \{A,B\}$, $\bm{p}_{f,i}= \begin{bmatrix} x_{f,i} & y_{f,i} & z_{f,i} \end{bmatrix}^{\intercal} \in \mathbb{R}^3$\linebreak denotes the foot position, and $\bm{l_{i}} = \bm{p_{c}}-\bm{p}_{f,i} \in \mathbb{R}^3$ represents the vector from the foot to the point mass; let $l_{0}>0$ be the rest length of this vector, which is the same for both legs. In contrast to the regular model, the spring stiffness $k_{i}>0$ can now be configured with different values for each leg $i$, while point masses were added to the feet of the model. Let $m_{f,i}$ be the mass of the foot in leg $i$, concentrated at the end point, which comes into contact with the ground. Finally, the swing leg touchdown is determined by the forward and lateral touchdown angles $\theta \in \mathbb{R}$ and $\phi \in \mathbb{R}$, respectively.

 According to~\cite{vasilopoulos2014compliant}, a compliant surface can be modeled using a combination of lumped parameter elements, based on viscoelastic theory. In this work, the \acf{HC} model will be utilized, since it is simple and fairly accurate \cite{stronge2018impact}. The \ac{HC} model captures the compliance of the surface through the interaction force between the materials that come into contact (e.g. foot and ground). Specifically, the interaction force applied at the foot of leg $i$ is defined as:
\begin{equation}
    F_{g,i} =  k_{g,i}\left(-z_{f,i}\right)^{h}-b_{g,i}\dot{z}_{f,i}\left(z_{f,i}\right)^{h},
\label{eq:18}
\end{equation}
where $k_{g,i}, b_{g,i} \in \mathbb{R}$ are the stiffness and damping parameters of the surface under the foot-leg $i$, respectively, $h=1.5$ for a Hertzian non-adhesive contact, and \linebreak$z_{f,i} \in \mathbb{R}$ is the vertical position of the foot in leg $i$. The damping of the surface is defined as a function of the stiffness: $b_{g,i} = 1.5 c_{a} k_{g,i}$, where $c_{a}$ will be fixed to 0.2, as in \cite{vasilopoulos2014compliant}. Consequently, by adjusting the stiffness of the ground $k_{g,i}$, the interaction force can be derived, as a function of the foot's vertical position $z_{f,i}$, which when negative represents penetration into the ground. Following the same notation as in \cite{vasilopoulos2014compliant}, the interaction force is assumed to act only on the vertical axis, while the foot motion is assumed to be constrained in the horizontal plane, meaning that the foot masses are only allowed to move vertically $(\dot{x}_{f,B} = \dot{y}_{f,B}\equiv 0)$.

Therefore, in contrast to the rigid case, in compliant surfaces the vertical position of the feet is allowed to change throughout the motion. Hence, this property has to be accounted for in the dynamics and the state of the system. During walking, the 3D \ac{Dual-SLIP} alternates between \ac{SS} and \acf{DS} phases. In \ac{SS}, the motion of the system is governed by the dynamics
\begin{align}
\begin{split}
    m\bm{\ddot{p}_{c}} &= \bm{F_{s,i}}+m\bm{g},\\
    m_{f,i}\bm{\ddot{p}_{f,i}} &=  F_{g,i}\bm{\hat{z}}-\bm{F_{s,i}} + m_{f,i}\bm{g},\\
    \bm{F_{s,i}} &=  k_{i}\left(l_{0}-\|\bm{l_{i}}\|\right)\bm{\hat{l}_{i}},
\end{split}
\label{eq:21}
\end{align}
where $\bm{F_{s,i}} \in \mathbb{R}^3$ is the spring force from leg $i$, $\bm{\hat{l}_{i}}$ is the unit vector along the leg in support $i$, and \linebreak$\bm{g} = \begin{bmatrix} 0& 0& -9.81\end{bmatrix}^{\intercal} \in \mathbb{R}^3$ is the gravity acceleration vector.

As in~\cite{liu2015dynamic}, it is assumed that the swing leg and the corresponding foot mass do not affect the dynamics of the system. Moreover, it is assumed that the swing leg's touchdown leg length is equal to the rest length $l_{0}$, while the vertical position and velocity of its foot mass are zero. In \ac{DS}, the motion of the system is governed by the dynamics\linebreak
\vspace*{-1em}
\begin{align}
\begin{split}
    m\bm{\ddot{p}_{c}} &= \bm{F_{s,A}} + \bm{F_{s,B}}+m\bm{g},\\
    m_{f,A}\bm{\ddot{p}_{f,A}} &=  F_{g,A}\bm{\hat{z}}-\bm{F_{s,A}} + m_{f,A}\bm{g},\\
    m_{f,B}\bm{\ddot{p}_{f,B}} &=  F_{g,B}\bm{\hat{z}}-\bm{F_{s,B}} + m_{f,B}\bm{g},
\end{split}
\label{eq:22}
\end{align}
where $\bm{F_{s,A}}$ and $\bm{F_{s,B}}$ are the spring forces from leg $A$ and $B$, respectively. 

\begin{figure}[h!]
    \centering
    \includegraphics[width =0.45\textwidth]{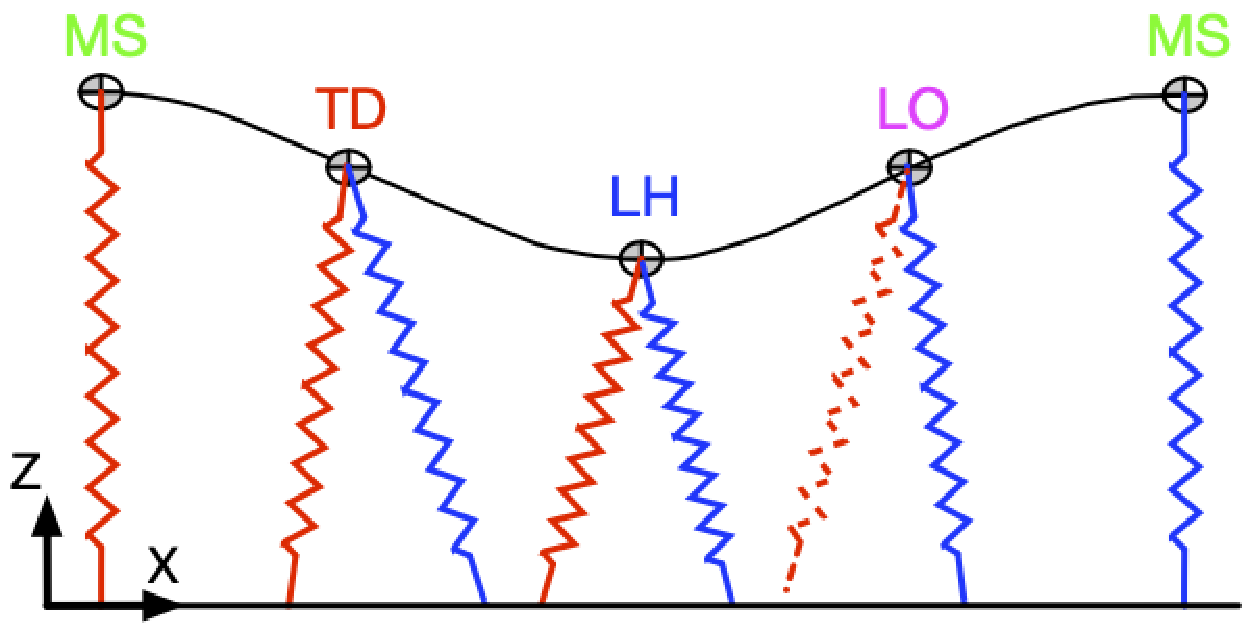}
    \caption{Sagittal plane view of nominal ``human-like” \ac{CoM} trajectory in a full walking step of the 3D \ac{Dual-SLIP} model.}
    \label{fig:gait_events}
\end{figure}

Throughout this paper, a walking step will be defined as the interval between two subsequent \ac{MS} gait events. During one step starting from \ac{MS}, four distinct gait events take place. Initially, \ac{MS} happens during the \ac{SS} phase when $\dot{z}_{c}=0$. Then, the swing leg touches down at \ac{TD} and the system enters the \ac{DS} phase. Next, the \ac{CoM} reaches its lowest height at \ac{LH}, and finally the leg originally in support lifts off at \ac{LO}. At \ac{LO}, the system reenters a \ac{SS} phase and the step is completed with the next \ac{MS} event, as shown in Fig.\ref{fig:gait_events}.

Each gait event can be associated with a gait event surface, where each event takes place when the \ac{CoM} state $\left(\bm{\dot{p}}_{c}, \bm{p}_{c}\right)$ crosses the corresponding surface. For a step where a leg $A$ is initially in support, the following surfaces are defined:
\begin{align}
    S_{MS}=&\left\{\left(\bm{\dot{p}}_{c}, \bm{p}_{c}\right)\big\vert \; \dot{z}_{c}=0\;,\; z_{c}>z_{TH}, \|\bm{l_{A}}\|<l_{0}\right\},\label{eq:3}\\
    S_{TD}=&\left\{\left(\bm{\dot{p}}_{c}, \bm{p}_{c}\right)\big\vert \; \dot{z}_{c}<0\;,\; z_{c}=z_{TH},  \|\bm{l_{A}}\|\leq l_{0}\right\},\label{eq:4}\\
    S_{LH}=&\left\{\left(\bm{\dot{p}}_{c}, \bm{p}_{c}\right)\big\vert \; \dot{z}_{c}=0\;,\; z_{c}<z_{TH},  \|\bm{l_{A}}\|\leq l_{0}\right\},\label{eq:5}\\
    S_{LO}=&\left\{\left(\bm{\dot{p}}_{c}, \bm{p}_{c}\right)\big\vert \; \dot{z}_{c}>0\;,\;   \|\bm{l_{A}}\|=l_{0}\right\},\label{eq:6}
\end{align}
where $z_{TH} = l_{0}\sin{\theta}$ is the threshold \ac{CoM} height at which \ac{TD} takes place. More importantly, the surfaces $S_{TD}$ and $S_{LO}$ are the switching surfaces for the hybrid dynamics of the system, as they determine when the system should switch from \ac{SS} to \ac{DS} dynamics, and vice versa. 
%%%%%%%%%%%%%%%%%%%%%%%%%%%%%%%%%%%%%%%%%%%%%%%%%%%%%%%%%%%%%%%%%%%%%%%%%%%%%%%%%%%%%%%%%%%%%%%%%%%%%%%%%%%%%%%%%%%
\subsubsection{Finding Periodic Gaits}
\label{subsection_periodic_gaits}
As in~\cite{liu2015dynamic}, we employ a nonlinear optimization approach to obtain suitable values for the state and control variables that lead to periodic, left-right symmetric walking gaits. Now, this optimization method relies on the symmetry of the \ac{CoM} about \ac{LH} over one step. In the case of uneven terrain, however, that symmetry ceases to exist \cite{liu2015trajectory}. As a result, the half-step optimization can only be utilized for locomotion over terrains with high stiffness values.

To derive the stride map, we consider a slice of the full 3D \ac{Dual-SLIP} state associated with the \ac{MS} event; i.e.,
\begin{equation}
    \bm{x} = [x_{c}-x_{f,i} \quad y_{c}-y_{f,i} \quad z_{c} \quad \dot{x}_{c} \quad \dot{y}_{c}]^\intercal,
\label{eq:7}
\end{equation}
where $i \in \{A,B\}$ again denotes the leg in support. Moreover, the forward and lateral touchdown angles together with the leg stiffness value, same for both legs, will be considered as control inputs available for regulating the state evolution of the system; that is,
$\bm{u} = [\theta \quad \phi \quad k]^\intercal$, where $k_{A}=k_{B}=k$.
%\label{eq:8}
%\end{equation}
%where $\theta$ is the forward touchdown angle, $\phi$ is the lateral touchdown angle and $k$ is the common stiffness of the support legs .

Following the notation of~\cite{liu2015dynamic}, let $\bm{x}_{n}$ and $\bm{u}_{n}$ denote the values of the state and control variables at the $n$-th \ac{MS} event. Then, the state at the next \ac{MS} event can be computed as $\bm{x}_{n+1}=\bm{f}(\bm{x}_{n},\bm{u}_{n})$, where the map $\bm{f}$ is calculated numerically by integrating the dynamics according to the sequence of events shown in Fig.~\ref{fig:gait_events}. Note here that the states $\bm{x}_{n}$ and $\bm{x}_{n+1}$ refer to \ac{SS} phases with different legs providing support; e.g., if $\bm{x}_{n}$ refers to the $n$-th \ac{MS} with leg $A$ providing support, the state $\bm{x}_{n+1}$ refers to the $(n+1)$-th\linebreak \ac{MS} with leg $B$ providing support. In this work, we will be concerned with nominal walking gaits that are periodic and left-right symmetric. This implies that---nominally---the states $\bm{x}_{n}$ and $\bm{x}_{n+1}$ corresponding to subsequent \ac{MS} events must be related by the following symmetry condition
\begin{equation}
\label{eq:sym}
   \bm{x}_{n+1} = \bm{A} \bm{x}_{n}
\end{equation}
where $\bm{A} = diag(1,-1,1,1,-1)$. In words, this condition implies that in the walking gaits we consider here, the forward and vertical position and velocity remain constant from one step to the next while the lateral position and velocity alternate their sign.

%To complete a 2-step gait,  the state $\bm{x}_{n+1}$ is used as initial condition of the next step and with inputs $\bm{u}_{n+1}$ the dynamics of the 3D \ac{Dual-SLIP} are integrated forward until the next \ac{MS} event to obtain $\bm{x}_{n+2}$. 
%Following the same notation as in \cite{liu2015dynamic}, an MS state return map $\bm{f}$ can be derived as: $\bm{x}_{n+1}=\bm{f}(\bm{x}_{n},\bm{u}_{n})$, where $\bm{x}_{n}$ and $\bm{u}_{n}$ are the \ac{MS} state and control input variables at step $n$. For a 2-step periodic \C{and left-right symmetric} locomotion, $\bm{x}^{A}_{n+2}=\bm{A}\bm{x}^{A}_{n}$ should be satisfied, where $\bm{A} = diag(1,-1,1,1,-1)$. Given the definition of the \ac{MS} state, this condition describes a 1-step motion with constant forward position and velocity, constant vertical position, and sign-alternating lateral position and velocity. 

To obtain suitable values for the control input and state variables that lead to a periodic left-right symmetric gait, we adopt the quarter-period (from the \ac{MS} to the \ac{LH} event) nonlinear optimization method suggested in~\cite{liu2015dynamic}. In more detail, the objective of the optimization is to ensure that the projection of the \ac{CoM} on the ground at the \ac{LH} event lies directly between the two support feet. Assuming---without loss of generality---that leg $A$ provides support, this can be achieved by minimizing the index\footnote{As shown in \cite{liu2015dynamic}, minimizing \eqref{eq:9} is a sufficient condition to achieve 2-step periodic, left-right symmetric gaits.}
%~\cite{liu2015dynamic} proposed a quarter-period (from \ac{MS} to \ac{LH}) nonlinear optimization method. Specifically, they formulated the following optimization problem to enforce that the ground projection of the \ac{CoM} at \ac{LH} is directly between the two support feet: with a desired forward velocity at \ac{MS}, 
\begin{align}
\begin{split}
    &\min_{\bm{u}_{0},z_{0}} \left\{ \left\Vert \frac{1}{2}\left( x_{f,A}+x_{f,B}\left(\bm{x_{0}},\bm{u_{0}}\right) \right) -x_{c}\left(t_{LH};\bm{x_{0}},\bm{u_{0}}\right) \right\Vert^{2} \right. \\
     &+ \left. \left\Vert \frac{1}{2}\left( y_{f,A}+y_{f,B}\left(\bm{x_{0}},\bm{u_{0}}\right) \right) -y_{c}\left(t_{LH};\bm{x_{0}},\bm{u_{0}}\right) \right\Vert^{2} \right\},
\label{eq:9}
\end{split}
\end{align}
subject to the dynamics of the system initiated at \ac{MS}; in \eqref{eq:9}, $t_{LH}$ is the time instance where the first \ac{LH} takes place and $\bm{x_{0}}$, $\bm{u_{0}}$ denote the initial \ac{MS} state and control input variables, respectively. As in~\cite{liu2015dynamic}, we restrict the optimization search for $\bm{x_{0}}$ to the following family of states
\begin{align}
\begin{split}
    \bm{x}_{0} &= [x_{0,d} \quad y_{0,d} \quad z_{0} \quad \dot{x}_{0,d} \quad \dot{y}_{0,d}]\\ 
    x_{0,d} &= 0\;m,\; y_{0,d} = 0.05 \; m,\; \dot{y}_{0,d} = 0 \; m/s 
\label{eq:10}
\end{split}
\end{align}
where $\dot{y}_{0,d} = 0 \; m/s$ is needed to satisfy the periodic gait conditions in~\cite[Equation (9)]{liu2015dynamic} and for a rest leg length $l_{0} = 1 \; m$, we set $y_{0,d}=0.05 \; m$, similar to~\cite{liu2015dynamic}. Regarding the remaining two state variables, the forward velocity $\dot{x}_{0,d}$ at \ac{MS} is specified by the user and the height $z_{0}$ is a decision variable. The optimizer then selects values for $z_{0}$ and the input variables $\bm{u}_0$ to minimize the cost function \eqref{eq:9}.     

%Furthermore, their optimization search was constrained to the following family of \ac{MS} state variables:
%Regarding the remaining two state variables,  $z_{0}$ and $\dot{x}_{0,d}$, the latter one is the desired forward velocity at $MS$, which is up to the user to select. The former one is an optimization variable, meaning that it is tuned along with the control variables, in order to minimize the cost function in question (Eq. (\ref{eq:9})). 
%Therefore, for a desired forward velocity, the optimizer will derive the optimal set of values for the initial height of the \ac{CoM} ($z_{0}$) and the control variables ($\bm{u}_{0}$), which initiate a 2-step periodic gait. 

In this work, we first implemented the proposed method for the regular 3D \ac{Dual-SLIP}, using a forward velocity of 1 $m/s$, $m=80 \; kg$, $l_{0}=1 \; m$, and derived the following optimal set of parameters using the nonlinear least-squares function \textit{lsqnonlin} in Matlab\textsuperscript{TM}:
\begin{align}
    \begin{split}
        \bm{x}_{0} &= [x_{0,d} \quad y_{0,d} \quad z_{0} \quad \dot{x}_{0,d} \quad \dot{y}_{0,d}] \\
        &= [0 \; m \quad 0.05 \; m \quad 0.99 \; m \quad 1 \; m/s \quad 0 \; m/s],%0.987530775175059
    \end{split}\label{eq:11}\\
    \begin{split}
        \bm{u}_{0} &= [\theta_{0} \quad \phi_{0} \quad k_{0} ]^\intercal \\
        &= [107.26^{\circ} \quad 10.94^{\circ} \quad 14163.54 \; N/m].%14163.5345312576
    \end{split}
\label{eq:12}
\end{align}
%where the desired forward velocity and the optimal set of values for $z_{0}$ and $\bm{u}_{0}$ are shown in bold.
%72.7411042049505 0.190971419146386
%%%%%%%%%%%%%%%%%%%%%%%%%%%%%%%%%%%%%%%%%%%%%%%%%%%%%%%%%%%%%%%%%%%%%%%%%%%%%%%%%%%%%%%%%%%%%%%%%%%%%%%%%%%%%%%%%%%

\subsubsection{The \ac{LQR} Controller}
\label{subsection_LQR_rigid}

Let $\bm{x}_{0}^{*},\bm{u}_{0}^{*}$ be an optimal set of parameters resulting in a left-right symmetric gait for a specific forward velocity. 
Along this gait, the \ac{MS} states evolve according to \eqref{eq:sym} so that the \emph{nominal} $n$-th \ac{MS} state is given by $\bm{x}_{n}^{*} = \bm{A}^{n}\bm{x}_{0}^{*}$ under the condition that the control parameters are selected according to $\bm{u}_{n}^{*} = \bm{B}^{n}\bm{u}_{0}^{*}$ with $\bm{B} = diag(-1,1,1)$ to account for the sign-alternating forward touchdown angle at each step. Under non-nominal conditions, however, initiating the system with  $\bm{x}_{0}^{*}, \bm{u}_{0}^{*}$ will not result in periodic locomotion due to the presence of disturbances. Thus, if $\bm{x}_{n}$ denotes the \emph{actual} value of state at the $n$-th \ac{MS} event, we have $\bm{x}_{n} \ne \bm{x}_{n}^{*}$. To ensure that the actual \ac{MS} state $\bm{x}_{n}$ approaches the nominal periodic evolution $\bm{x}_{n}^{*}$, a discrete-time, infinite-horizon \ac{LQR} will be designed; the procedure closely follows~\cite{liu2015dynamic}, and thus our exposition here will be terse.
%Therefore, the actual \ac{MS} state has to be feedback-regulated to achieve periodic locomotion. For this purpose, a discrete-time infinite-horizon \ac{LQR} controller was utilized in \cite{liu2015dynamic}.

Let $\Delta \bm{x}_{n}=(\bm{x}_{n}-\bm{x}_{n}^{*})$, $\Delta \bm{u}_{n}=(\bm{u}_{n}-\bm{u}_{n}^{*})$, \linebreak$\Delta \tilde{\bm{x}}_{n}=\bm{A}^{n}\Delta \bm{x}_{n}$, $\Delta \tilde{\bm{u}}_{n}=\bm{B}^{n}\Delta \bm{u}_{n}$. In \cite{liu2015dynamic}, it is shown that 
\begin{equation}
    \Delta \tilde{\bm{x}}_{n+1} \approx \bm{J}_{x}\Delta \tilde{\bm{x}}_{n} + \bm{J}_{u}\Delta \tilde{\bm{u}}_{n},
\label{eq:13}
\end{equation}
where $\bm{J}_{x}=\bm{A}\frac{\delta \bm{f}}{\delta \bm{x}}$ and $\bm{J}_{u} = \bm{A}\frac{\delta \bm{f}}{\delta \bm{u}}$ are evaluated at $(\bm{x}_{0}^{*},\bm{u}_{0}^{*})$. Now, consider the following quadratic cost for positive definite matrices $\bm{Q}$ and $\bm{R}$:
\begin{align}
    &\min_{\Delta \tilde{\bm{u}}} \sum_{n=0}^{\infty}\Delta \tilde{\bm{x}}_{n}^{T}\bm{Q} \Delta \tilde{\bm{x}}_{n} + \Delta \tilde{\bm{u}}_{n}^{T}\bm{R} \Delta\tilde{\bm{u}}_{n}\label{eq:14}\\
    &\text{s.t. } \Delta \tilde{\bm{x}}_{n+1} = \bm{J}_{x}\Delta\tilde{\bm{x}}_{n} + \bm{J}_{u}\Delta\tilde{\bm{u}}_{n}.
    \label{eq:15}
\end{align}
Then, if $\left(\bm{J}_{x},\bm{J}_{u}\right)$ is controllable, the following time-invariant feedback gain is obtained:
\begin{equation}
    \bm{K} = -(\bm{J}_{u}^{T}\bm{P}\bm{J}_{u} + \bm{R})^{-1}\bm{J}_{u}^{T}\bm{P}\bm{J}_{x},
\label{eq:16}
\end{equation}
where $\bm{P}$ is the unique solution of the Discrete-Time Algebraic Riccati Equation (DARE). This results in the following time-invariant control law
\begin{equation}
    \bm{u}_{n}=\bm{u}_{n}^*+\bm{B}^{n}\bm{K}\bm{A}^{n}(\bm{x}_{n}-\bm{x}_{n}^{*})
\label{eq:17}
\end{equation}
which adjusts the control input at each \ac{MS} event to regulate the state so that it converges to the target periodic gait. Note that in this work, we will refer to a controller obtained for $\bm{Q}=\bm{R}=\bm{I}$, where $\bm{I}$ is the identity matrix, as an identity \ac{LQR} controller. 

In order to verify the validity of the modified dynamics that account for the compliance of the terrain, locomotion over high stiffness values was tested to simulate rigid terrain walking. After running multiple simulations with different stiffness values, it was concluded that a stiffness of \linebreak$k_{g,A}=k_{g,B}=50  \; MN/m$ resulted in a response closest to the one observed for the ideal rigid case, described in previous works~\cite{liu2015dynamic,liu2015dual}. Specifically, the system was able to reach at least 100 steps, which can be interpreted as a sign of stable performance and hence successful implementation of the proposed methodology \cite{geyer2006compliant}.
Therefore, the stiffness of $50 \; MN/m$ will be considered from now on as the equivalent of the rigid terrain, meaning that any terrain with a stiffness lower than that will be considered as compliant. In the aforementioned simulations, the system was simulated using 
$m=80 \; kg$, $l_{0}=1 \; m$, and $m_{f,A/B}=1 \; kg$, while the same optimal initial conditions and identity \ac{LQR} controller were utilized, as the ones analyzed in Sections~\ref{subsection_periodic_gaits} and~\ref{subsection_LQR_rigid}.

%\begin{table}[t]
%    \centering
%    \begin{tabular}{!{\vrule width 1.5pt}c!{\vrule width 1.5pt}c|c|c|c!{\vrule width 1.5pt}}
%    \ChangeRT{1.5pt}
%        \textbf{Parameter} & \textbf{Value} & \textbf{Units} \\\ChangeRT{1.5pt}
%        Point mass $(m)$ & 80 & kg\\\hline
%        Foot Masses $(m_{f,A/B})$ & 1 & kg\\\hline
%        Leg Rest Length $(l_{0})$ & 1 & m\\\ChangeRT{1.5pt}
%    \end{tabular}
%    \caption{Model parameters of the 3D \ac{Dual-SLIP} in compliant terrain.}
%    \label{tab:model_param}
%\end{table}
%%%%%%%%%%%%%%%%%%%%%%%%%%%%%%%%%%%%%%%%%%%%%%%%%%%%%%%%%%%%%%%%%%%%%%%%%%%%%%%%%%%%%%%%%%%%%%%%%%%%%%%%%%%%%%%%%%%
\subsection{One-step Unilateral Low Stiffness Perturbations}
\label{subsection_one_step_perturbs}

\begin{figure}[h!]
    \centering
    \includegraphics[width=0.49\textwidth]{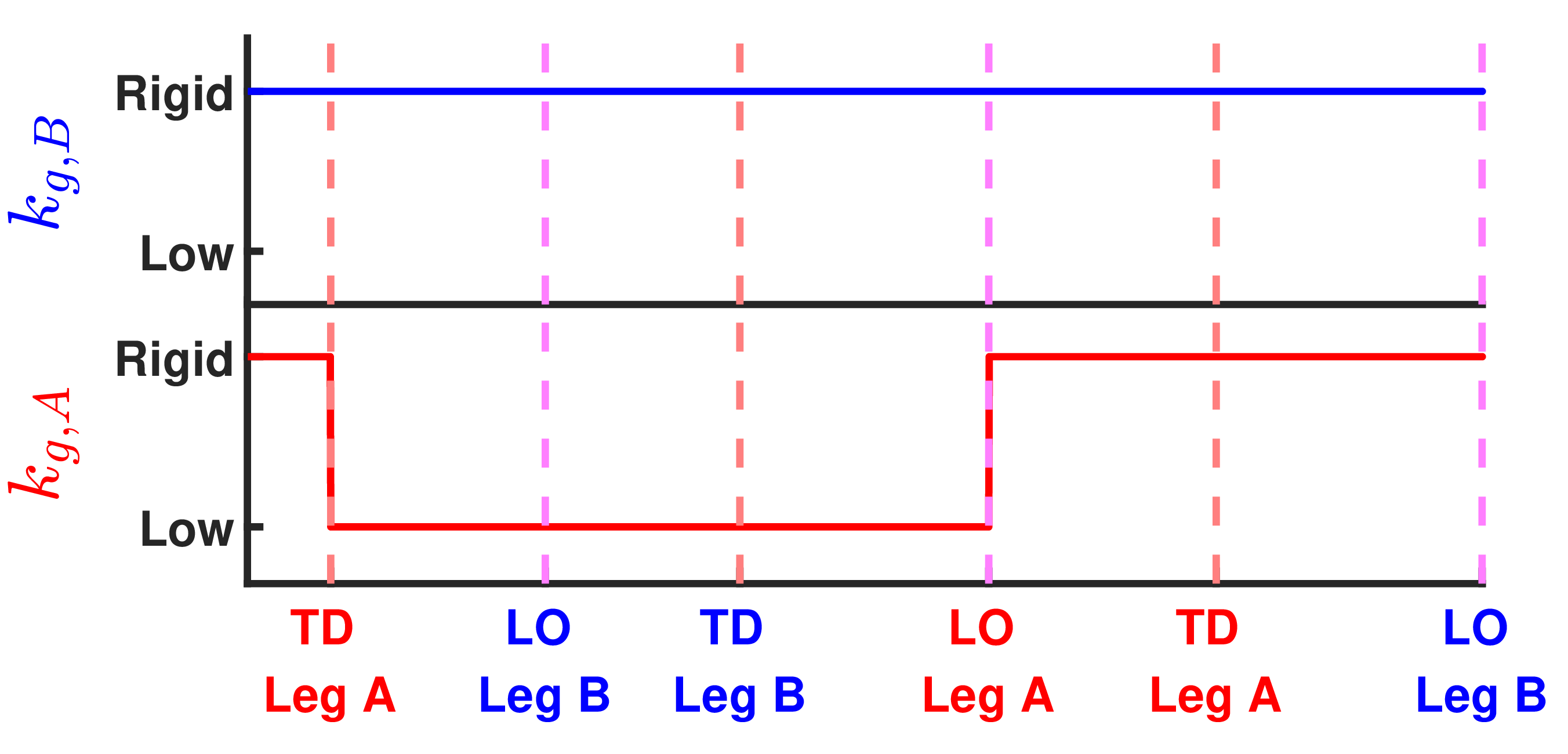}
    \vspace*{-1.8em}
    \caption{Timing of the one-step unilateral low stiffness perturbation. Top and bottom figures illustrate the ground stiffness values underneath the legs $B$ and $A$, respectively. Rigid ground stiffness corresponds to $50 \; MN/m$, while low ground stiffness can take any value lower than that. The color of the label for each gait event indicates the related leg (blue for leg $B$, red for leg $A$).}
    \label{fig:timing_perturbation}
\end{figure}

%why are we interested in these perturbations? Perhaps mention VST experiments 
As a first step towards achieving periodic gait over compliant terrains, we investigate the response of the model to one-step unilateral low stiffness perturbations. For these simulations, the model was initiated using a set of optimal parameters to achieve periodic gait, while the ground stiffness was set to the rigid value of $50 \; MN/m$. Then, after $n_{p}$ steps the system experienced a one-step unilateral lower stiffness perturbation, after which the ground stiffness was set back to rigid as depicted in Fig. \ref{fig:timing_perturbation}. Specifically, during the $n_{p}$ step the ground stiffness under the leg about to land ($A$) was lowered to a specific value at \ac{TD} and was kept constant throughout the whole stance phase of that leg (\ac{TD} to \ac{LO}). Then, the ground stiffness was reset to rigid for the rest of the trial. The ground stiffness of the other leg ($B$) remained fixed to rigid throughout the whole trial. It should be noted that such perturbations have been applied to humans before for understanding human gait and for rehabilitation purposes using a novel instrumented device \cite{skidmore2016unilateral,skidmore2014investigation,skidmore2014variable}.
\setlength{\textfloatsep}{10pt plus 2.0pt minus 4.0pt}

%%%%%%%%%%%%%%%%%%%%%%%%%%%%%%%%%%%%%%%%%%%%%%%%%%%%%%%%%%%%%%%%%%%%%%%%%%%%%%%%%%%%%%%%%%%%%%%%%%%%%%%%%%%%%%%%%%%
\subsection{Biomechanics-inspired Proposed Controller}
\label{subsection_proposed_controller}

Inspired by human locomotion, we propose a modified controller for the 3D \ac{Dual-SLIP}, which adjusts the stiffness of the legs in support, to withstand one-step unilateral low stiffness perturbations. 
Previously, Ferris et al. showed that runners increased leg stiffness for their first step when transitioning from hard to softer surfaces \cite{ferris1999runners}. As the runners were expecting the perturbation, they tended to pre-adjust the increased leg stiffness during their last step on the hard surface. Inspired by this, we propose a controller that increases the leg stiffness of both legs of the 3D \ac{Dual-SLIP} to handle expected one-step unilateral low stiffness perturbations.

\begin{figure}[h]
\centering

\hspace*{-0.8cm}\begin{tikzpicture}[scale=0.5, every node/.style={scale=0.5},  title/.style={font=\fontsize{12}{8}\color{black!90}\ttfamily}]
\large
    \node[anchor=south west,inner sep=0] at (0,0) {\hspace*{-1cm}\includegraphics[scale=0.65]{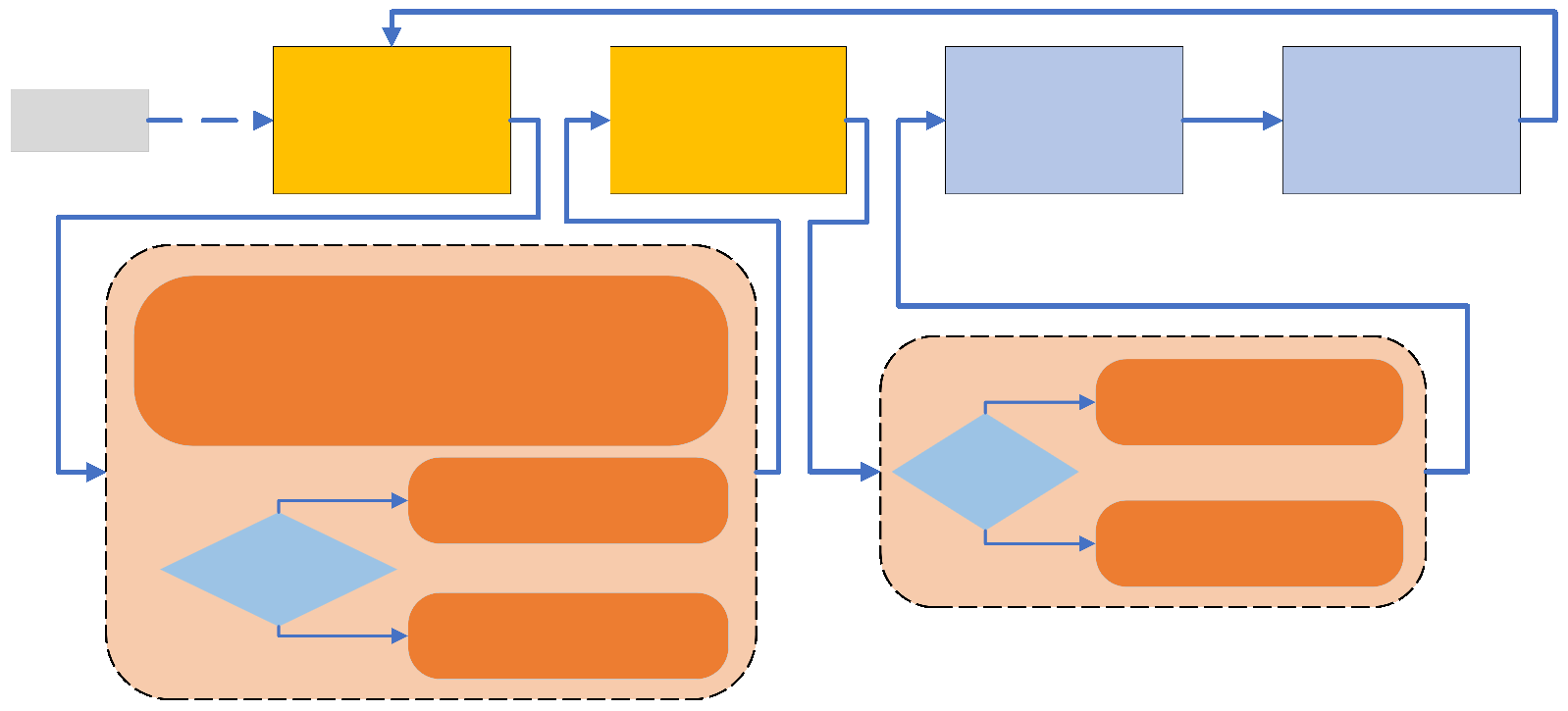}};
    \node[text width=3cm,align=center] at (-0.1,6.65) {Start};
    \node[text width=3cm,align=center] at (1.35,7) {$\bm{x_{0}^{*}}$};
    \node[text width=3cm,align=center] at (1.35,6.3) {$\bm{u_{0}^{*}}$};
    \node[text width=3cm,align=center] at (3.4,6.65) {\ac{SS} Dynamics until \ac{MS}};
    \node[text width=3cm,align=center] at (7.15,6.65) {\ac{SS} Dynamics until \ac{TD}};
    \node[text width=3cm,align=center] at (10.95,6.65) {\ac{DS} Dynamics until \ac{LH}};
    \node[text width=3cm,align=center] at (14.75,6.65) {\ac{DS} Dynamics until \ac{LO}};
    \node[text width=3cm,align=center] at (-0.9,4) {$n, \bm{x_{n}}$};
    \node[text width=2cm,align=center] at (2.1,1.57) {$n=n_{p}+1$};
    \node[text width=2cm,align=center] at (10,2.65) {$n=n_{p}$};
    % \node[text width=5.2cm,align=center] at (11.6,0.8) {\underline{Second Part}};
    \node[text width=4cm,align=center] at (3.85,4.5) {\underline{LQR Controller}};
    \node[text width=8cm,align=center] at (3.85,3.9) {$\bm{u}_{n}=\bm{u}_{n}^*+\bm{B}^{n}\bm{K}\bm{A}^{n}(\bm{x}_{n}-\bm{x}_{n}^{*})$};
    \node[text width=4cm,align=center] at (3.85,3.4) {$[\theta_{n} \quad \phi_{n} \quad k_{n} ]^\intercal = \bm{u}_{n}$};
    % \node[text width=2cm,align=center] at (3.45,3.35) {$\phi = u_{n}(2)$};
    % \node[text width=2cm,align=center] at (3.45,2.9) {$k = u_{n}(3)$};
    \node[text width=2cm,align=center] at (5.1,2.6) {$k_{B} = k_{n}$};
    \node[text width=2cm,align=center] at (5.1,2.1) {$k_{A} = k_{n}$};
    \node[text width=3cm,align=center] at (4.8,1.07) {$k_{B} = k_{n_{p}+1}$};
    \node[text width=3cm,align=center] at (5.0,0.6) {$k_{A} = k_{1}k_{n_{p}+1}$};
    \node[text width=3cm,align=center] at (12.7,3.7) {$k_{B} = k_{B}$};
    \node[text width=3cm,align=center] at (12.7,3.3) {$k_{A} = k_{A}$};
    \node[text width=3cm,align=center] at (12.6,2.05) {$k_{B} = k_{2}k_{n_{p}}$};
    \node[text width=3cm,align=center] at (12.6,1.65) {$k_{A} = k_{1}k_{n_{p}}$};
    \node[text width=3cm,align=center] at (10.7,1.65) {YES};
    \node[text width=3cm,align=center] at (10.7,3.7) {NO};
    \node[text width=3cm,align=center] at (2.8,0.55) {YES};
    \node[text width=3cm,align=center] at (2.8,2.65) {NO};
    %titles for boxes
    \node[text width=3cm,align=center] at (6.65,0.8) {(\textbf{Q3})};
    \node[text width=3cm,align=center] at (6.65,2.3) {(\textbf{Q1})};
    \node[text width=3cm,align=center] at (14.25,1.85) {(\textbf{Q2})};
    \node[text width=3cm,align=center] at (14.25,3.45) {(\textbf{Q4})};
\end{tikzpicture}
\caption{Overview of the biomechanics-inspired proposed controller. The model is initiated with the optimal state-control pair ($\bm{x_{0}^{*}}, \bm{u_{0}^{*}}$) to achieve periodic gait with a desired forward velocity at \ac{MS}. The \{\ac{MS}, \ac{TD}\} and \{\ac{LH}, \ac{LO}\} gait events are identified using \acf{SS} and \acf{DS} dynamics, respectively. At every \ac{MS} event, the \ac{LQR} controller is implemented and the $\bm{u_{n}}$ feedback law is derived. During the perturbation step ($n=n_{p}$), the stiffness of both legs is initially determined by the \ac{LQR} controller at \ac{MS} (Q1) and then it is amplified at \ac{TD} by the control gains $k_{1}$ and $k_{2}$ (Q2). At the next step ($n=n_{p}+1$), the leg experiencing the perturbation ($A$) maintains an increased stiffness at \ac{MS}, while the stiffness of the leg about to land on rigid terrain ($B$) is determined based on the \ac{LQR} controller with no adjustment (Q3). Then, at \ac{TD}, the stiffness of both legs is not altered (Q4). Finally, the stiffness for both legs is determined based on the \ac{LQR} for all other steps.}
% \vspace*{-2em}
\label{fig:proposed_controller}
\end{figure}
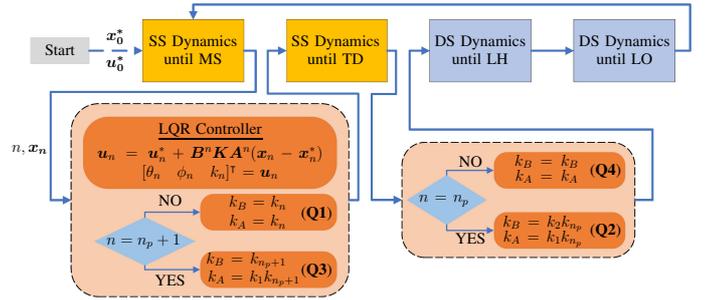

The proposed feedback law is based on the \ac{LQR} controller developed in Section~\ref{subsection_LQR_rigid}, modified to allow for further stiffening of the legs when needed; an overview of the proposed controller is illustrated in Fig.~\ref{fig:proposed_controller}.
%, which regulates the \ac{MS} state to achieve periodic locomotion using the optimal state-control pair ($\bm{x_{0}^{*}}, \bm{u_{0}^{*}}$). Therefore, the proposed controller can be thought of as a extended version of the baseline standard controller. 
In more detail, initially both legs share the same stiffness value, as determined by the \ac{LQR} controller at each step $n$; i.e., $k_{A}=k_{B}=k_{n}$. Then, at the \ac{TD} event of the perturbation step ($n_{p}$), the stiffness of the leg about to land is increased to $k_{A} = k_{1}k_{n_{p}}$, where $k_{1}>1$ is a control gain and $k_{n_{p}}$ is the stiffness value derived by the \ac{LQR} controller for that step. At the same time, the stiffness of the leg in support is also increased to $k_{B} = k_{2}k_{n_{p}}$, where $k_{2}>1$ is again a control gain. These stiffness values remain constant as long as each leg is in stance phase. At the \ac{MS} event of the following step ($n_{p}+1$), the stiffness of the leg experiencing the perturbation, retains the same control gain $k_{A} = k_{1}k_{n_{p}+1}$, while the stiffness of the leg about to land on rigid terrain is set back to $k_{B} = k_{n_{p}+1}$. Finally, by the time the next \ac{MS} event takes place, the leg that experienced the perturbation has switched to swing phase (\ac{LO}) and is about to land on rigid terrain. Therefore, from that point on, both legs share again the same stiffness value $k_{n}$, as it is calculated by the \ac{LQR} controller at each step.

%%%%%%%%%%%%%%%%%%%%%%%%%%%%%%%%%%%%%%%%%%%%%%%%%%%%%%%%%%%%%%%%%%%%%%%%%%%%%%%%%%%%%%%%%%%%%%%%%%%%%%%%%%%%%%%%%%%

%% file: results.tex
\section{Results}
In this section, the proposed controller will be compared to the standard \ac{LQR} controller designed for rigid surfaces, with respect to their response to unilateral one-step low stiffness perturbations. For all simulations, the model parameters mentioned in the Methods Section were used and the model was initiated with the optimal set of parameters ($\bm{x_{0}^{*}}, \bm{u_{0}^{*}}$) shown in \eqref{eq:11}-\eqref{eq:12}. For all one-step unilateral low stiffness perturbations, the perturbation took place at the tenth step ($n_{p}=10$). All simulations were implemented and executed in MATLAB\textsuperscript{TM} version 9.7 (R2019b), where the nonlinear least-squares function \texttt{lsqnonlin} and the embedded fixed-step integrator \texttt{ode4} were utilized for the optimization and the dynamic simulation, respectively.

\subsection{Performance of the Standard \ac{LQR} Controller}
\begin{figure}[t]
% \vspace*{-0.5cm}
\centering
\hspace*{-0.5cm}    \includegraphics[scale=0.27]{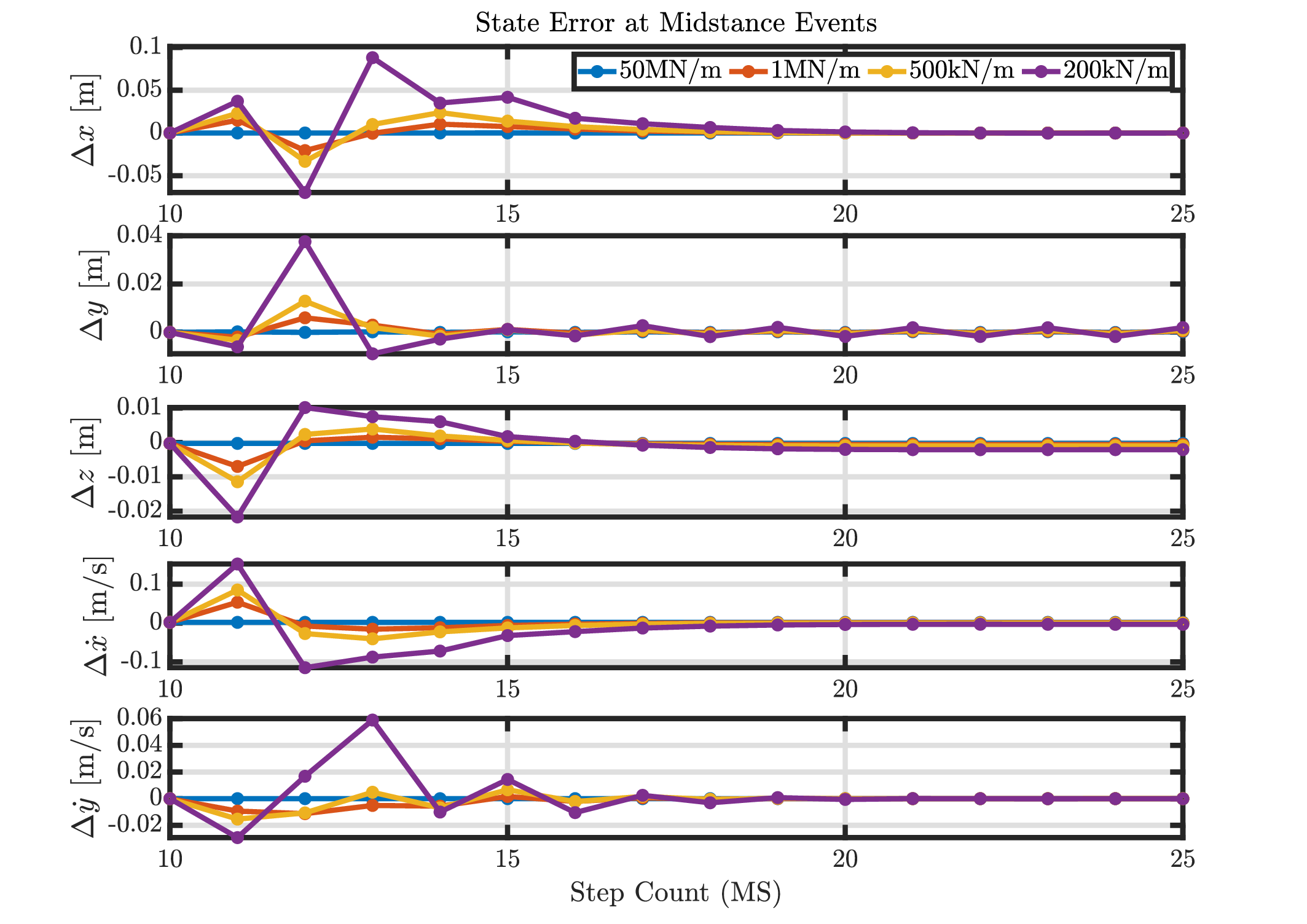}%   \caption{1b}
\vspace*{-0.2cm}
    \caption{State error response for stiffness perturbations of 50 $MN/m$, 1 $MN/m$, 500 $kN/m$ and 200 $kN/m$ using the standard \ac{LQR} controller.}
    \label{fig:state_input_error_50_10_1_500_200kN/m}
\end{figure}

% Rigid controller
Initially, the robustness of the standard \ac{LQR} controller was explored under one-step unilateral perturbations of various ground stiffness levels. In all simulations, an identity \ac{LQR} controller was used. For perturbation ground stiffness values ranging from 50 $MN/m$ to 200 $kN/m$, the system was shown to be able to endure the one-step perturbation and reach the threshold performance of 100 steps. Four representative cases are shown in Fig. \ref{fig:state_input_error_50_10_1_500_200kN/m} for ground stiffness values of $50 \; MN/m$, $1 \; MN/m$, $500 \; kN/m$ and $200 \; kN/m$. It should be noted that in all four cases the model was able to achieve the desired number of 100 steps, but for brevity we chose to show the system response only for up to the $25^{th}$ step. As it can be observed, the perturbation introduces errors in all state variables, the magnitude of which increases as the perturbation stiffness decreases. Nevertheless, the \ac{LQR} controller is able to regulate the introduced errors and lead the system to steady-state for all cases. Although the error is minimized in less than 10 steps, small steady-state errors are evident for some state variables, which again increase as the perturbation stiffness decreases.

For ground stiffness values lower than 200 $kN/m$, the perturbation destabilized the system and caused it to fail, i.e. the system was not able to complete a proper step after the perturbation. Considering the stiffness levels reported in \cite{vasilopoulos2014compliant}, the 200 $kN/m$ stiffness level would be classified as moderate ground. As a result, it appears that the standard \ac{LQR} controller proposed for locomotion over rigid terrain is able to handle one-step unilateral low stiffness perturbations, only up to moderate ground stiffness values. 

\vspace*{-0.1cm}
\subsection{Performance of the Proposed Controller}
\begin{figure}[!t]
% \vspace*{-0.5cm}
% \begin{subfigure}{0.48\textwidth}
  \centering
\hspace*{-0.5cm}    \includegraphics[scale=0.27]{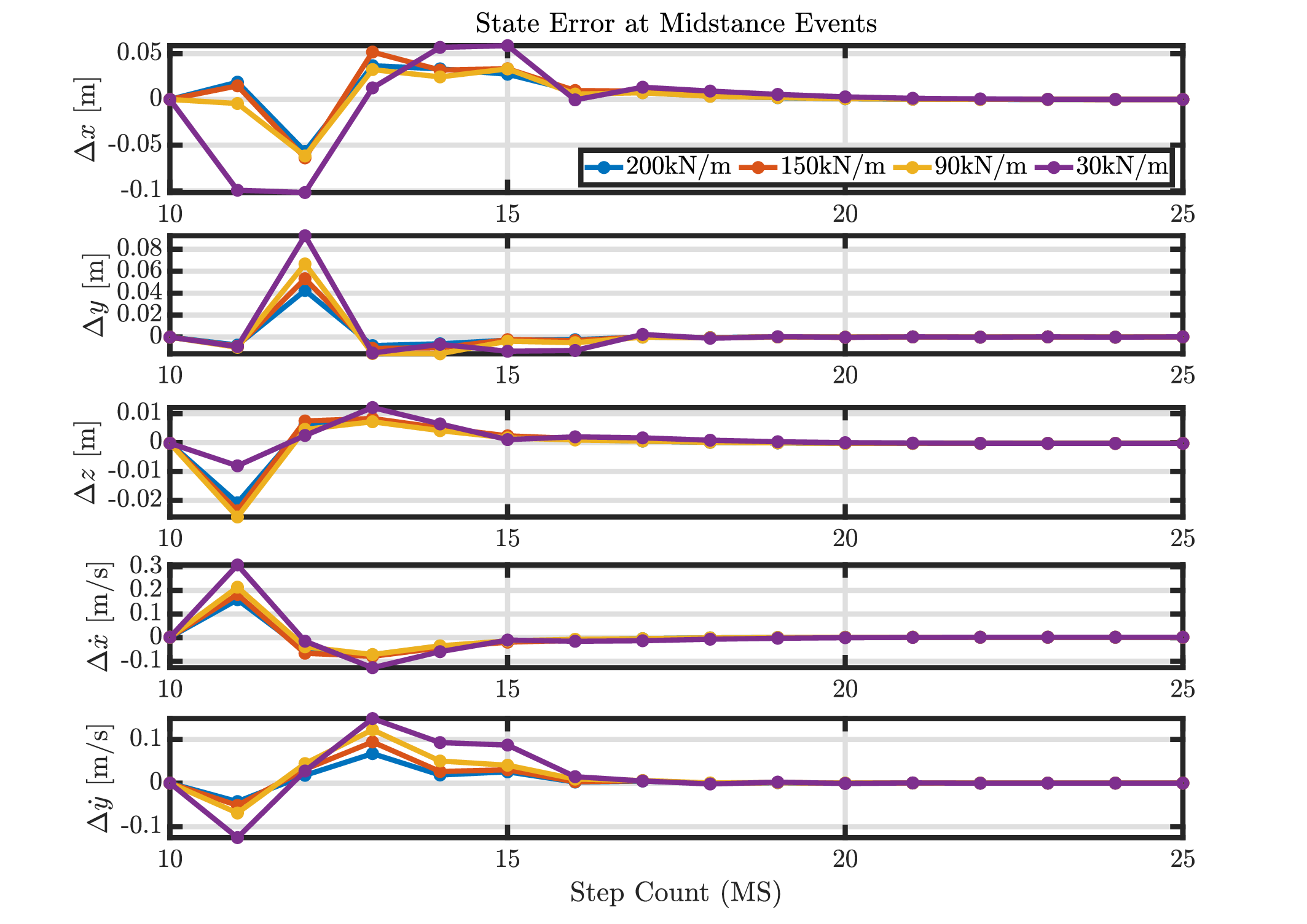}
\vspace*{-0.65cm}
    \caption{State error response for stiffness perturbations of 200, 150, 90 and 30$\;kN/m$ using the proposed controller.}
    \label{fig:state_input_error_200_150_90_30kN/m}
\end{figure}

% Proposed controller
As an extension of the standard \ac{LQR} controller, the proposed controller inherits its stable performance for perturbation ground stiffness values ranging from 50 $MN/m$ to 200 $kN/m$. Therefore, we focus only on stiffness values lower than the 200 $kN/m$ threshold. Again, identity $\bm{Q}$ and $\bm{R}$ matrices were utilized for the \ac{LQR} part of the proposed controller. For stiffness values lower than 200 $kN/m$, we showed that the system with the proposed controller is able to endure perturbations of stiffness as low as 30 $kN/m$. It should be noted that this stiffness value is much lower than the \textit{soft ground} category of 80 $kN/m$ reported in \cite{vasilopoulos2014compliant}, and it resembles walking on a
foam pad \cite{bosworth2016robot}. Four representative cases are shown for one-step perturbations of 200, 150, 90 and 30 $kN/m$ in Fig.~\ref{fig:state_input_error_200_150_90_30kN/m}. Similarly to Fig. \ref{fig:state_input_error_50_10_1_500_200kN/m}, we chose to show the system response only up to the $25^{th}$ step, although the desired number of 100 steps was achieved for all cases.  

Similar to the higher stiffness levels shown in Fig.~\ref{fig:state_input_error_50_10_1_500_200kN/m}, the perturbations again introduce errors, the magnitude of which increases as the ground stiffness decreases. As it can be seen in all cases, the model manages to handle the perturbation taking place at the $10^{th}$ step, while the proposed controller regulates any introduced errors and leads to zero steady-state errors. The rapid recovery of the system is to be noted, as the error is suppressed in less than 10 steps. Moreover, by comparing the responses of the model for the 200 $kN/m$ perturbation between the standard and the proposed controller, it is clear that the proposed controller leads to smaller errors during both the transient and the steady-state response. For all four perturbation stiffness values, the control gains ($k_{1},k_{2}$) were tuned to minimize the steady-state errors. The control gains used are listed in Table \ref{tab:control_gains}, where it can be seen that as the perturbation stiffness decreases, higher control gains have to be used to handle the perturbation and achieve zero steady-state errors. 

\begin{figure*}[t!]
    \centering
    \includegraphics[width=0.94\textwidth]{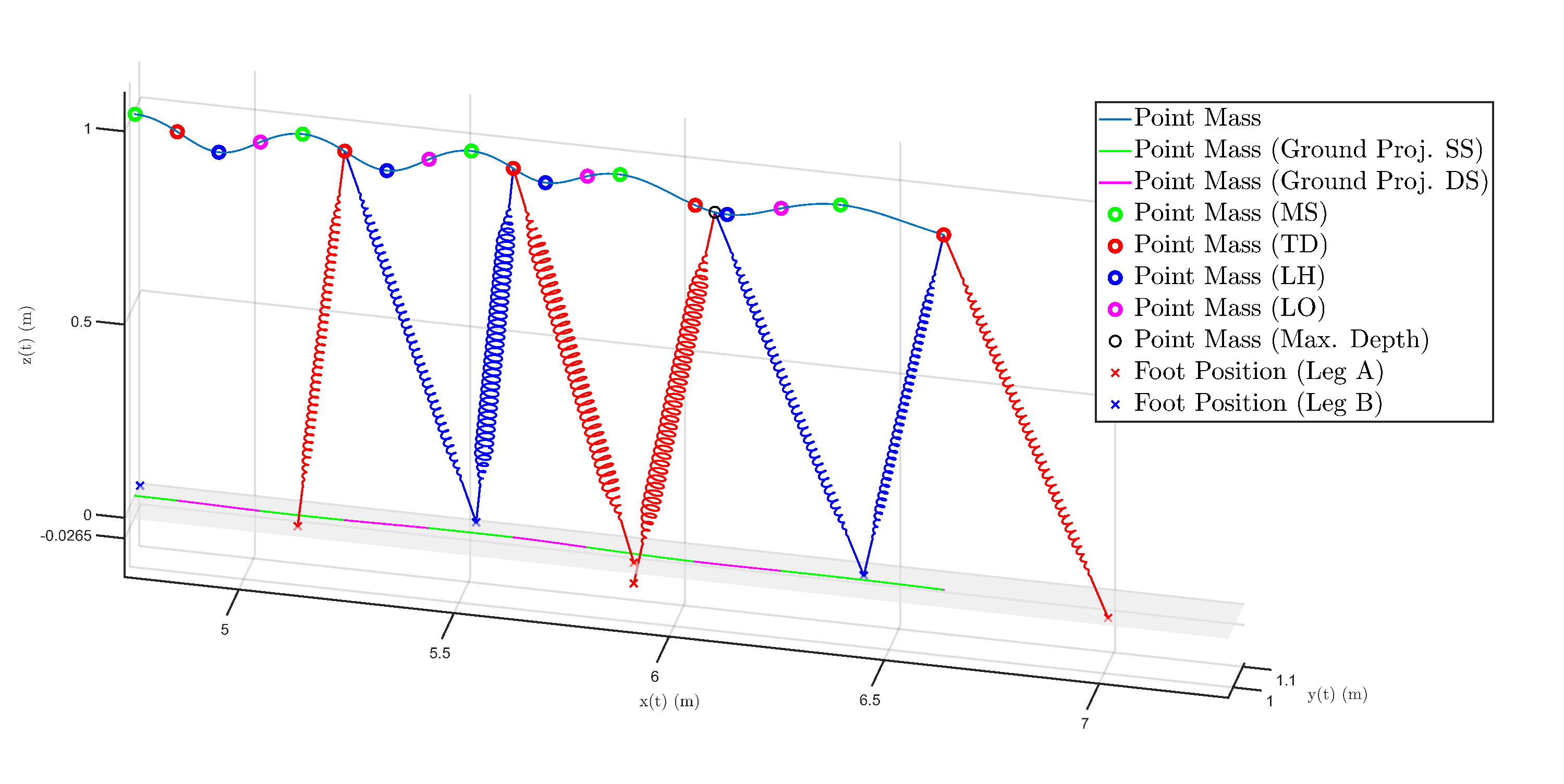}
    \vspace*{-0.8cm}
    \caption{The 3D Dual-SLIP model experiencing an one-step unilateral stiffness perturbation of 200 $kN/m$. Blue line on top illustrates the three dimensional trajectory of the \ac{CoM}, while the green and magenta lines on the bottom denote its projection on the x-y plane, during the \ac{SS} and \ac{DS} phases, respectively. Green, red, blue and magenta circles ($\circ$) represent the position of the \ac{CoM} during the \ac{MS}, \ac{TD}, \ac{LH} and \ac{LO} gait events, respectively. Black circle indicates the position of the \ac{CoM} when the perturbed foot reaches the maximum penetration depth (0.0265 $m$). Red and blue crosses ($\times$) depict the position of the feet for legs $A$ and $B$, respectively. The increased coil radius of the spring legs in the second and third snapshot of the model indicates an increase in leg stiffness due to the proposed controller.\vspace*{-0.2cm}}
    \label{fig:simulation}
\end{figure*}

Figure \ref{fig:simulation} shows the model experiencing a stiffness perturbation of 200 $kN/m$, while a video demonstration of the 3D Dual-SLIP experiencing one-step unilateral stiffness perturbations in simulation can be found at \cite{simulation_video}. Before the perturbation, the stiffness of the legs is set based on the internal \ac{LQR} controller. At the \ac{TD} event during the perturbation step, the stiffness of the legs is amplified throughout each leg's stance phase. Then, during the perturbation, the perturbed leg reaches the maximum foot penetration depth, maintaining the amplified stiffness, while the stiffness for the unperturbed leg is again set based on the \ac{LQR} controller. Finally, after the perturbation, both legs share again the same stiffness, as calculated by the \ac{LQR} controller.

In order to highlight the significance and the physical meaning of the perturbations, the maximum penetration depth of the foot stepping on the soft surface is provided in Table \ref{tab:control_gains} for all four cases. As expected, lower perturbation stiffness values correspond to deeper penetration depths. More importantly, given that the leg rest length is 1 $m$, the model manages to regulate an extensive vertical sinking of the perturbed leg, close to 12\% of the leg's rest length, in the case of the lowest stiffness of 30 $kN/m$.

\begin{table}[!t]
    \centering
    \begin{tabular}{!{\vrule width 1.5pt}c!{\vrule width 1pt}c!{\vrule width 1.5pt}c|c|c|c!{\vrule width 1.5pt}}
    \ChangeRT{1.5pt}
        \multicolumn{2}{!{\vrule width 1.5pt}c!{\vrule width 1.5pt}}{\textbf{Perturbation Stiffness (kN/m)}} & 200 & 150 & 90 & 30\\\ChangeRT{1.5pt}
        \multirow{2}{*}{\textbf{Control Gains}} & $k_{1}$ & 1.05 & 1.1 & 1.4 & 8.4 \\\cline{2-6}
         & $k_{2}$ & 1.14 & 1.19 & 1.36 & 3.17\\\ChangeRT{1.5pt}%1.1422 & 1.1923 & 1.36 & 3.1705
        \multicolumn{2}{!{\vrule width 1.5pt}c!{\vrule width 1.5pt}}{\textbf{Max Penetration Depth (cm)}} & 2.65 & 3.21 & 4.64 & 11.49\\\ChangeRT{1.5pt}
    \end{tabular}
    \caption{Control gains and maximum penetration depth for one-step unilateral stiffness perturbations using the proposed controller.}
    \label{tab:control_gains}
    \vspace*{-0cm}
\end{table}
% Compare gains to the ones observed in human walking

%%%%%%%%%%%%%%%%%%%%%%%%%%%%%%%%%%%%%%%%%%%%%%%%%%%%%%%%%%%%%%%%%%%%%%%%%%%%%%%%%%%%%%%%%%%%%%%%%%%%%%%%%%%%%%%%%%%
%%%%%%%%%%%%%%%%%%%%%%%%%%%%%%%%%%%%%%%%%%%%%%%%%
%%%%%%%%%%%%%%%%%%%%%%%%%%%%%%%%%%%%%%%%%%%%%%%%%

%% file: conclusions.tex
\section{Conclusion}

This paper extends the 3D Dual-SLIP model to support for the first time locomotion over compliant terrains and proposes a novel biomechanics-inspired controller to regulate one-step unilateral low stiffness perturbations. Using a standard \ac{LQR} controller, the extended model is shown to be able to endure such perturbations only up to a moderate ground stiffness level of 200 $kN/m$. On the contrary, the proposed controller can produce stable gait at stiffness levels as low as 30 $kN/m$, which results in vertical sinking of the $1m$-long leg as deep as 11.49 $cm$. Therefore, the proposed controller allows for robust dynamic walking over extremely low stiffness one-step unilateral perturbations. As robust and stable walking over a wide range of compliant terrains is an important problem for legged locomotion, this work can significantly advance the field of bipedal walking by improving the control of bipeds and humanoids, as well as prosthetic devices with tunable stiffness.